\documentclass[letterpaper]{article} 
\usepackage[preprint]{aaai2027}              
\usepackage[hyphens]{url}  
\usepackage{graphicx} 
\usepackage{natbib}  
\usepackage{caption} 
\usepackage{algorithm}
\usepackage{algorithmic}

\usepackage{newfloat}
\usepackage{listings}
\DeclareCaptionStyle{ruled}{labelfont=normalfont,labelsep=colon,strut=off} 
\floatstyle{ruled}
\newfloat{listing}{tb}{lst}{}
\floatname{listing}{Listing}

\usepackage{booktabs}
\usepackage{multirow} 
\usepackage{amssymb} 

\title{Rendering-in-the-Loop: An Execution-Driven Agent for Interactive Web Development}
\author{
    Yilong Guo\textsuperscript{\rm 1}\equalcontrib,
    Hanqi Chen\textsuperscript{\rm 1}\equalcontrib,
    Zixiao Ye\textsuperscript{\rm 2},
    Guanzhong Wang\textsuperscript{\rm 1}\corresponding,
    Chen Yu\textsuperscript{\rm 2},
    Zeyu Chen\textsuperscript{\rm 1}
}
\affiliations{
    \textsuperscript{\rm 1}Baidu Inc., Beijing, China\\
    \textsuperscript{\rm 2}School of Computer Science and Technology, Huazhong University of Science and Technology, Wuhan, China\\
    \{guoyilong, chenhanqi, wangguanzhong\}@baidu.com
}

\begin{document}

\maketitle

\begin{abstract}

Multimodal large language models have achieved remarkable progress in front-end web development, generating interactive webpages from multimodal references such as screenshots and interaction videos. However, existing work largely emphasizes visual metrics such as aesthetics and layout similarity, while overlooking the more critical validation of interactive functionality. We present \textbf{RILA}, an execution-driven agent that puts browser rendering in the loop, iteratively editing generated code from runtime interaction feedback. RILA introduces an \textbf{Action Interaction Verification (AIV)} module that replays the reference interaction trajectory on the generated webpage to collect grounded execution-aware observations, and an \textbf{Execution-Aware Rendering Score (ERS)} that jointly measures interaction correctness and visual fidelity to guide iterative optimization. We further build an execution-verified data synthesis pipeline that produces diverse, high-quality training data, offering gains complementary to inference-time optimization. On IWR-Bench, RILA consistently improves both interaction and visual fidelity across foundation models. Notably, with our training pipeline, RILA lifts the compact Qwen3.5-9B backbone from \textbf{40.40\%} to \textbf{57.52\%}, surpassing far larger one-shot generators, including the 1T-parameter Kimi-K2.6 (\textbf{55.61\%}) and the proprietary GPT-5.5 (\textbf{55.74\%}).

\end{abstract}


\section{Introduction}

\begin{figure}[!t]
\centering
\includegraphics[width=\columnwidth]{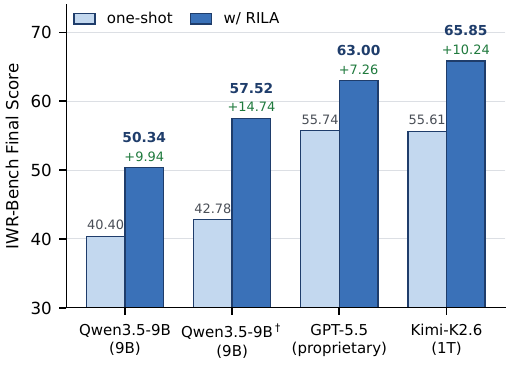}
\caption{Final scores on IWR-Bench. RILA improves every backbone over its one-shot output. Notably, the fine-tuned Qwen3.5-9B$^\dagger$ refined by RILA (\textbf{57.52}) surpasses every one-shot generator, including the proprietary GPT-5.5 (55.74) and the 1T-parameter Kimi-K2.6 (55.61). Green numbers denote the gain over one-shot generation.}
\label{fig1}
\end{figure}

Recent advances in multimodal large language models (MLLMs) have significantly advanced web development from diverse inputs, including textual instructions, screenshots, sketches~\citep{li2025sketch2code}, and interaction videos~\citep{xiao2024interaction2code,chen2025iwr}. Powered by strong coding models~\citep{hui2024qwen25coder,guo2024deepseekcoder}, modern systems can synthesize complete web applications consisting of HTML, CSS, JavaScript, and external assets with impressive visual quality, while coding agents further improve generation through iterative tool use and environment interaction~\citep{yang2024sweagent}. These advances have made automatic web development increasingly practical for interface prototyping and application development.

Despite this progress, a usable webpage must satisfy two distinct requirements, \emph{visual fidelity} (how closely the rendered page matches the reference appearance) and \emph{interaction correctness} (whether the page behaves correctly under user actions), which are often optimized in isolation, and improving one does not guarantee the other. Existing multimodal web development methods optimize static rendering quality through one-shot generation, and dedicated front-end repair approaches~\citep{yuan2024designrepair} audit that rendering against design guidelines; both produce pages that look plausible but break under user actions. Software engineering agents instead repair code from compiler errors, unit tests, or textual logs; lacking any rendering feedback, they can restore functional behavior yet leave the visual result poor. Either way, the decisive failures surface only during runtime interaction, such as broken event bindings, incorrect state transitions, failed resource loading, and unexpected rendering after user operations; such failures erode interaction correctness yet stay invisible to the static screenshots or textual logs on which these methods rely.

To address this, we propose \textbf{RILA} (\textbf{R}endering-\textbf{I}n-the-\textbf{L}oop \textbf{A}gent), an execution-driven agent that closes the loop between browser execution and code editing through runtime interaction feedback. At its core, an \textbf{Action Interaction Verification (AIV)} module runs each generated webpage in a real browser, replays the reference interaction trajectory, verifies every action, and records the rendering state after each step. From these observations, an \textbf{Execution-Aware Rendering Score (ERS)} jointly measures interaction correctness and visual fidelity, and is used to retain the historically best implementation so that unsuccessful edits do not cause degradation. In parallel, a multimodal critique model turns the runtime observations into structured fix suggestions. Guided by this critique, RILA localizes and repairs faulty code with software engineering (SWE) tools, iterating over execution--critique--edit cycles.

Beyond inference-time refinement, we build an execution-verified data synthesis pipeline that jointly samples webpage domains, interaction capabilities, and visual styles under compatibility constraints and validates every sample through browser execution, yielding diverse and reliable training data. On IWR-Bench, RILA consistently improves both interaction and visual fidelity across foundation models, and this data proves complementary. Even with only a compact critique model, RILA substantially improves the fine-tuned Qwen3.5-9B over one-shot generation; and under a stronger critique (Kimi-K2.6), this compact 9B model surpasses both the 1T-parameter Kimi-K2.6 and the proprietary GPT-5.5 that generate webpages in one shot (Figure~\ref{fig1}), indicating that execution-driven refinement with a strong critique can offset a large gap in code-model scale.

Our contributions are summarized as follows:

\begin{itemize}
    \item We propose \textbf{RILA}, an execution-driven agent that puts browser rendering in the loop, refining webpages through an execution--critique--edit cycle driven by runtime interaction feedback rather than static screenshots or textual logs.

    \item We turn browser execution into a verifiable optimization signal via an \textbf{Action Interaction Verification (AIV)} module that replays reference trajectories and checks each action, and an \textbf{Execution-Aware Rendering Score (ERS)} that measures interaction correctness and visual fidelity to guide critique generation and best-implementation selection.

    \item We propose an \textbf{execution-verified} data synthesis pipeline that samples from a compatibility-constrained taxonomy of domains, interaction capabilities, and visual styles and retains only browser-verified pages. Extensive experiments on IWR-Bench show that RILA consistently improves webpage generation quality across backbone models, with this data providing further complementary gains.
\end{itemize}

\section{Related Work}
\subsection{Web Development}

Web development with large models is organized by reference modality. From text alone, WebGen-Bench asks an agent to lay out a runnable multi-file codebase and verifies each functionality by driving the deployed site with a navigation agent~\citep{lu2026webgen}. The visual setting maps an image to code: pix2code first mapped a GUI screenshot to markup~\citep{beltramelli2017pix2code}, WebSight synthesized screenshot--HTML pairs at scale for training~\citep{laurencon2024websight}, Design2Code established automatic rendering-similarity metrics on real webpages~\citep{si2024design2code}, Web2Code paired instruction-tuning data with webpage-understanding questions~\citep{yun2024web2code}, and Sketch2Code accepts rough sketches, resolving their ambiguity by querying a simulated designer~\citep{li2025sketch2code}.

Because monolithic generation omits and misarranges elements on complex layouts, later work adds structure. DCGen divides a screenshot into segments, generates code per segment, and reassembles them into a page~\citep{wan2024dcgen}; ScreenCoder splits the task across specialized grounding, planning, and generation agents, and reuses that pipeline as a data engine for fine-tuning~\citep{jiang2025screencoder}; WAFFLE instead specializes the model itself, with structure-aware attention over the HTML hierarchy and contrastive training that aligns UI images with code~\citep{liang2024waffle}.

A closer line makes generation iterative. UI2Code$^N$ embeds generation in a loop of execution and visual inspection, and trains on it with a preference objective that ranks rendered candidates against one another~\citep{yang2025ui2coden}; VisRefiner ties rendered-versus-target discrepancies to the code edits causing them and self-refines with reinforcement learning~\citep{deng2026visrefiner}; DesignRepair audits an interface against a Material Design knowledge base along two streams, its code and its rendered page~\citep{yuan2024designrepair}. All three read the render rather than exercise it. Interaction is meanwhile treated as an evaluation target, from interactive prototypes~\citep{xiao2024interaction2code} and authored interaction paths~\citep{wu2026interactive} to full user interaction videos~\citep{chen2025iwr}.

\begin{figure*}[t]
\centering
\includegraphics[width=\textwidth]{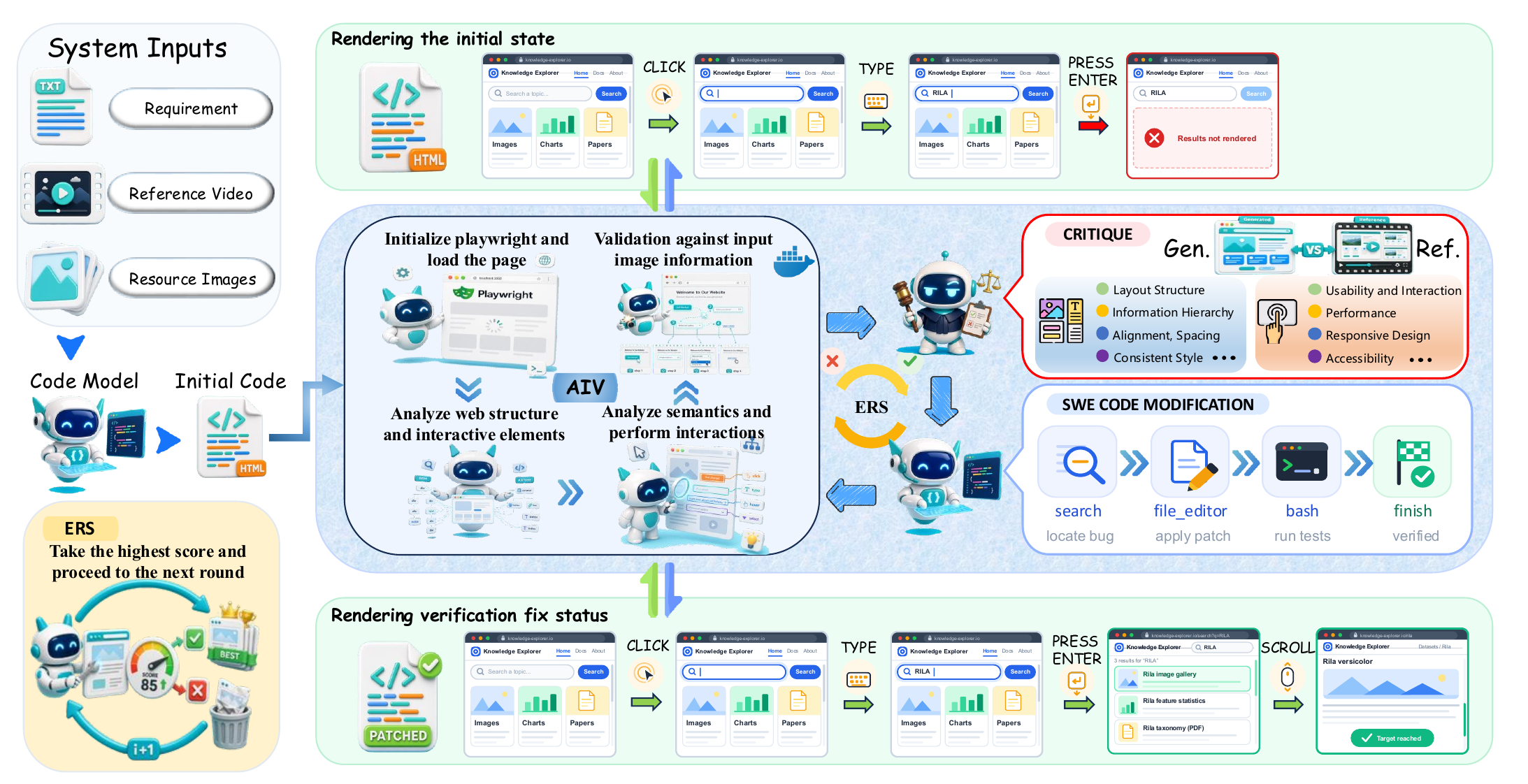}
\caption{
\textbf{Overview of RILA.}
RILA closes the loop between browser execution and code editing for interactive webpage generation.
Starting from an initial implementation generated by a code model, the proposed \textbf{Action Interaction Verification (AIV)} module executes the webpage in a real browser and collects runtime observations.
A critique model analyzes the execution results and visual discrepancies to generate fix suggestions, while the proposed \textbf{Execution-Aware Rendering Score (ERS)} maintains the best implementation across iterations to avoid performance degradation.
Guided by runtime observations and structured critique, the code model iteratively refines the webpage implementation, progressively improving rendering quality and interaction correctness.
}
\label{pipeline}
\end{figure*}

\subsection{Software Engineering Agents}

LLM-based software engineering agents resolve programming tasks by interacting with a development environment. Much design effort concerns the agent--codebase interface: SWE-agent tunes commands and feedback formats to what a model can reliably use rather than to human ergonomics~\citep{yang2024sweagent}, CodeAct makes executable code the action space, so actions compose and revise from interpreter output~\citep{wang2024codeact}, and OpenHands consolidates these into an open platform~\citep{wang2024openhands}. For defect resolution, AutoCodeRover navigates program structure with AST-aware search and fault localization~\citep{zhang2024autocoderover}, Agentless shows a fixed localize--repair--validate pipeline can rival agentic loops~\citep{xia2024agentless}, and AgentCoder pairs a programmer with test-design and test-execution agents that iterate until the tests pass~\citep{huang2023agentcoder}---progress measured by execution, on repository test suites~\citep{jimenez2024swebench} or in procedurally built verifier environments~\citep{jain2025r2egym}.

Web agents instead act inside live environments, perceiving rendered pages to operate real websites~\citep{zhou2023webarena,deng2023mind2web,zheng2024seeact,lu2024weblinx,xu2024agenttrek}, without authoring the code behind them. RILA brings execution inside the generation loop: it replays the reference interactions, verifies each action in a real browser, and drives code search and editing with feedback that static renders and textual logs cannot expose.

\section{Method}
\subsection{Overview}

RILA formulates reliable web development as an execution-driven incremental editing problem rather than one-shot generation or regenerate-from-scratch refinement. This design is motivated by two observations: many interaction failures surface only when the webpage is actually executed and are thus invisible to static code inspection or screenshot comparison, and regenerating the entire page after each observed failure tends to overwrite already-correct code, causing regressions and unstable optimization. RILA instead executes the webpage in a real browser, diagnoses the observed failures, and applies localized edits.

As illustrated in Figure~\ref{pipeline}, given a multimodal task specification
\(
x=\{r,v,s\}
\),
where $r$ denotes the textual instruction, $v$ is the reference interaction video, and $s$ denotes the accompanying resource images (e.g., logos and media assets), a code generation model $G(\cdot)$ first produces an initial webpage implementation $C_0 = G(x)$, the generated front-end project. From the reference video we further derive the reference demonstration $R$, comprising the reference screenshots and the interaction trajectory $A=\{a_1,\dots,a_N\}$ to be reproduced; $R$ serves as the target for both execution and evaluation.

RILA then iteratively optimizes the generated implementation through four tightly coupled components, detailed in the following subsections: \textbf{Action Interaction Verification (AIV)} executes the webpage with Playwright and replays the reference trajectory $A$ to collect runtime screenshots and action execution logs; \textbf{Execution-Aware Rendering Score (ERS)} scores interaction correctness and visual fidelity from these observations and retains the historically best implementation; \textbf{Execution-Aware Critique} turns the observed discrepancies against $R$ into structured fix suggestions; and \textbf{Tool-Assisted Code Editing} uses software engineering tools to localize and incrementally repair faulty code. The same execution-based verification further acts as a quality filter in our \textbf{Execution-Verified Data Synthesis} pipeline.


\subsection{Action Interaction Verification}

To evaluate the generated webpage under realistic conditions, the Action Interaction Verification (AIV) module executes the generated project in a real browser using Playwright, instead of relying on static screenshot comparison. Given a candidate implementation, AIV first loads the page and analyzes its rendered structure and interactive elements to ground each reference action to a concrete target. It then sequentially replays the reference interaction trajectory $A=\{a_1,\dots,a_N\}$, where each action $a_i$ corresponds to a browser operation such as clicking, typing, or scrolling, interpreting the surrounding page semantics to perform the interaction as a real user would.

During execution, Playwright records the status of each action and captures the rendered webpage after every successful interaction. Once an action fails, subsequent actions are skipped, faithfully simulating real usage where later steps become unreachable once an earlier one breaks. The resulting observation $O_t$ aggregates the captured screenshots and action execution logs. Note that $A$ serves only as the replay script for AIV; it is never exposed to the code generation model, which sees only the instruction and the reference screenshots and must therefore infer the intended interactions from appearance alone.

%
%


\subsection{Execution-Aware Rendering Score}

To quantitatively evaluate each implementation, RILA introduces an \textbf{Execution-Aware Rendering Score (ERS)} that jointly measures interaction correctness and visual fidelity under real browser execution. Computed directly from the AIV observations---the per-action verification results and the rendered screenshots captured during trajectory replay---ERS reflects how the webpage actually behaves at runtime rather than its static appearance, and serves as the criterion for selecting which implementation to edit next.

The overall rendering score is defined as

\begin{equation}
\mathrm{ERS}(C_t)
=
\frac{1}{|\mathcal{T}|}\sum_{S\in\mathcal{T}} S,
\quad
\mathcal{T}=\{S_{\mathrm{int}},\,S_{\mathrm{vis}}\},
\end{equation}

where $S_{\mathrm{int}}$ measures interaction correctness and $S_{\mathrm{vis}}$ measures visual similarity to the reference.

The interaction term is the execution success rate of the predefined action sequence, $S_{\mathrm{int}}=N_{\mathrm{success}}/N$, where $N$ is the total number of expected interaction steps and $N_{\mathrm{success}}$ denotes the number of actions verified as successful by AIV.

The visual term $S_{\mathrm{vis}}$ assesses rendering quality from multiple perspectives by aggregating a set of complementary visual metrics $\mathcal{M}$ that capture pixel-level similarity, textual consistency, and semantic structure similarity,

\begin{equation}
S_{\mathrm{vis}}
=
\frac{1}{|\mathcal{M}|}\sum_{S\in\mathcal{M}} S,
\quad
\mathcal{M}=\{S_{\mathrm{SSIM}},\,S_{\mathrm{OCR}},\,S_{\mathrm{SEM}}\},
\end{equation}

where $S_{\mathrm{SSIM}}$ is the Structural Similarity Index (SSIM) between rendered and reference screenshots, $S_{\mathrm{OCR}}$ measures OCR text similarity, and $S_{\mathrm{SEM}}$ denotes the cosine similarity between image embeddings extracted by a lightweight vision encoder.

Rather than always editing the most recent candidate, RILA edits the historically best implementation. After each round, it computes the ERS of the current candidate and updates the best implementation by $C^*=\arg\max_{C_i,\,i\le t}\mathrm{ERS}(C_i)$. The subsequent editing round is then initialized from $C^*$ rather than the latest webpage, so that unsuccessful edits do not degrade performance. Note that ERS only decides which implementation to carry forward; the editing itself is driven by the structured critique feedback described next.

%
%
%
%
%
%
%
%

\begin{figure*}[t]
\centering
\includegraphics[width=0.97\textwidth]{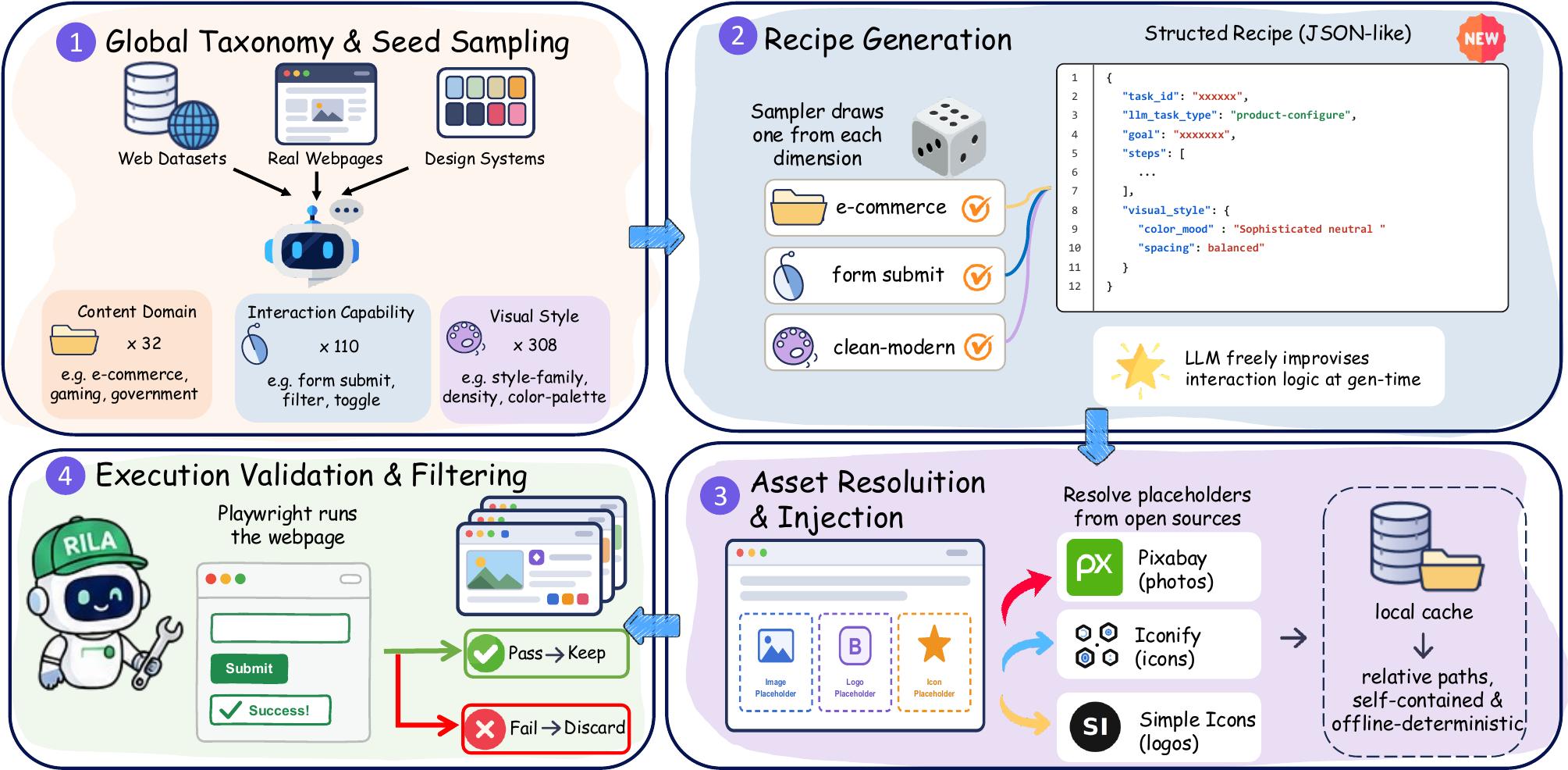}
\caption{
Our execution-verified data synthesis pipeline. Seeds sampled from a global taxonomy of domains, interaction capabilities, and visual styles are expanded into generation recipes, realized as complete webpages with real assets resolved and injected, and finally filtered by browser execution, which retains only pages whose every intended interaction succeeds.
}
\label{fig:data_pipeline}
\end{figure*}

\subsection{Execution-Aware Critique}

Given the execution observation $O^*$ of the current best implementation $C^*$ and the reference demonstration $R$, a multimodal critique model analyzes the discrepancies between the generated webpage and the expected behavior to produce critique feedback $\Delta_t=\mathcal{C}(O^*,R)$. The comparison spans both interaction and rendering aspects: it examines functional usability, such as whether each action triggers the expected behavior, as well as visual and design quality, including layout structure, information hierarchy, alignment and spacing, style consistency, responsive design, and accessibility.

Rather than emitting a full rewrite, the critique model summarizes the execution results into structured fix instructions, including the identified issue, supporting execution evidence, and actionable suggestions. Such execution-aware feedback transforms low-level browser observations into high-level editing objectives.


\subsection{Tool-Assisted Code Editing}

Given the structured critique $\Delta_t$, RILA performs incremental code editing with a tool-assisted SWE module $\mathcal{S}$, producing the round-$t$ candidate implementation $C_t=\mathcal{S}(C^*,\Delta_t)$.

This module incrementally edits the webpage by invoking software engineering tools such as semantic code search, file editing, shell execution, and runtime verification within an execution-based environment. Through multiple rounds of tool interaction, it progressively localizes faulty implementations, generates code patches, and validates the edited webpage.

The updated implementation is subsequently re-executed by the AIV module, forming a closed-loop optimization process that runs for $K$ rounds.



\subsection{Execution-Verified Data Synthesis}
\label{sec:data}
Beyond inference-time refinement, we further construct a large-scale training corpus of interactive webpages. Given a sampled task query, a powerful LLM synthesizes a complete webpage, which we then execute and retain only if all its intended interactions run successfully; this \textbf{execution-verified} pipeline yields diverse, functionally valid training data at scale. As illustrated in Figure~\ref{fig:data_pipeline}, it runs in four stages. We first sample seeds from a global taxonomy distilled from web-interaction datasets~\citep{deng2023mind2web,zheng2024seeact,lu2024weblinx,xu2024agenttrek}, screenshots, and design systems, spanning 32 content domains, 110 interaction capabilities, and 308 visual styles, with an LLM-built compatibility matrix filtering implausible combinations. A sampler then expands each compatible combination into a structured generation recipe, leaving the concrete interaction logic for the LLM to improvise at generation time. Typed image placeholders are next filled with assets from open-source libraries, cached locally and rewritten to relative paths so every page is self-contained and renders deterministically offline. Finally, each page is verified by execution: we replay its full interaction sequence and keep only pages on which every action succeeds, which rejects roughly one-third of the seeds and acts as a quality gate. The retained data is diverse across all three dimensions, with long-tailed visual styles (the top-20 cover only 18\%) that we find important for generalization.

\section{Experiments}

\begin{table*}[t]
\centering
\small
\setlength{\tabcolsep}{10pt}
\begin{tabular}{l c c c c c c c}
\toprule
Code Model & IFS $\uparrow$ & OCR $\uparrow$ & DINO $\uparrow$ & LFS $\uparrow$ & HFS $\uparrow$ & VFS $\uparrow$ & Final $\uparrow$ \\
\midrule
\multicolumn{8}{l}{\textit{w/o RILA} (one-shot generation)}\\
Kimi-K2.6            & 50.85 & 56.11 & 80.65 & 68.38 & 65.03 & 66.70 & 55.61 \\
GPT-5.5             & 50.12 & 61.49 & 84.05 & 72.77 & 64.99 & 68.88 & 55.74 \\
Qwen3.5-9B          & 34.28 & 48.63 & 73.48 & 61.06 & 48.34 & 54.70 & 40.40 \\
Qwen3.5-9B$^\dagger$      & 36.05 & 53.79 & 72.78 & 63.28 & 53.69 & 58.49 & 42.78 \\
\midrule
\multicolumn{8}{l}{\textit{w/ RILA} (Kimi-K2.6 critique)}\\
Kimi-K2.6          & 63.39 & 62.76 & 83.28 & 73.02 & 70.12 & 71.57 & 65.85 (\textbf{+10.24}) \\
GPT-5.5             & 58.24 & 67.28 & 87.44 & 77.36 & 70.81 & 74.08 & 63.00 (\textbf{+7.26}) \\
Qwen3.5-9B          & 46.90 & 54.15 & 76.52 & 65.33 & 51.43 & 58.38 & 50.34 (\textbf{+9.94}) \\
Qwen3.5-9B$^\dagger$      & 53.91 & 60.79 & 80.57 & 70.68 & 61.24 & 65.96 & 57.52 (\textbf{+14.74}) \\
\bottomrule
\end{tabular}
\caption{Main results on IWR-Bench. Each model is evaluated under one-shot generation (\textit{w/o RILA}) and under RILA (\textit{w/ RILA}), where RILA uses Kimi-K2.6 as the critique model. Parenthesized values report the Final-score improvement over the same model's one-shot baseline. $^\dagger$ denotes Qwen3.5-9B fine-tuned on our synthetic data.}
\label{tab:main_results}
\end{table*}

\subsection{Experimental Setup}

\noindent\textbf{Benchmark.}
We evaluate RILA on \textbf{IWR-Bench}~\citep{chen2025iwr}, a benchmark for interactive webpage reconstruction from multimodal inputs. It contains 113 tasks from 100 real-world websites; each provides a textual instruction, a reference interaction video, and the corresponding assets, and requires generating a self-contained HTML webpage that faithfully reproduces both the visual appearance and interactive behaviors of the reference.

\noindent\textbf{Models.}
We evaluate both proprietary and open-source foundation models, including Kimi-K2.6, GPT-5.5, and Qwen3.5-9B. In addition, we evaluate our supervised fine-tuned model, denoted as Qwen3.5-9B$^\dagger$, obtained by fine-tuning Qwen3.5-9B on our execution-verified synthetic data, to investigate the complementary effects of training-time adaptation and inference-time execution-driven refinement. During evaluation, different models can be assigned as the \emph{Critique Model} and the \emph{Code Model}. Without RILA, the code model directly generates webpages in a one-shot manner. With RILA, the critique model analyzes runtime execution results and provides structured fix suggestions to guide iterative code refinement. Unless otherwise specified, RILA uses Kimi-K2.6 as the critique model.

\noindent\textbf{Evaluation Metrics.}
Following IWR-Bench, all generated webpages are evaluated using its agent-as-a-judge protocol in a real browser environment. We report \textbf{Interactive Functionality Score (IFS)}, which measures the success rate of executing the reference action sequence, and \textbf{Visual Fidelity Score (VFS)}, which evaluates rendering quality from both low-level visual consistency and high-level semantic alignment. Specifically, VFS consists of the \textbf{Low-level Fidelity Score (LFS)}, computed from OCR similarity and DINO feature similarity, and the \textbf{High-level Fidelity Score (HFS)}, assessed by a frontier multimodal judge model. The benchmark further reports a \textbf{Final Score}, computed as the weighted combination of IFS and VFS, as the overall evaluation metric.

\noindent\textbf{Implementation Details.}
Unless otherwise specified, RILA performs three rounds of iterative refinement, with webpages executed in a headless Chromium browser via Playwright. The \textbf{Tool-Assisted Code Editing} module is built on the R2E-Gym execution environment~\citep{jain2025r2egym}, and each editing round runs for at most 50 tool-use steps. Unless otherwise stated, all experiments use identical prompts and hyperparameters across different backbone models.

\noindent\textbf{Training Details.}
We fine-tune Qwen3.5-9B on the execution-verified synthetic data using the Megatron backend of the ms-swift framework~\citep{zhao2025swift} on 16 H800 GPUs. We perform full-parameter fine-tuning of the language backbone while keeping the vision encoder and the vision--language aligner frozen. To accommodate complete front-end projects, we set the maximum sequence length to 65,536 tokens and cap the input image resolution at 921,600 pixels, and compute the loss only on the final response turn.

\subsection{Main Results}
Table~\ref{tab:main_results} reports the performance of different backbone models with and without RILA on IWR-Bench. To our knowledge, no existing refinement method reports results on IWR-Bench, so we compare each backbone against its own one-shot generation and, in the ablation below, against alternative refinement strategies instantiated within RILA. RILA consistently improves webpage generation quality across all evaluated models and evaluation dimensions, demonstrating that execution-driven iterative optimization is effective regardless of the underlying code generation capability. For the 9B backbone, directly applying RILA raises the final score from \textbf{40.40\%} to \textbf{50.34\%} (\textbf{+9.94}) without any additional training, simultaneously improving functional correctness and visual consistency.

Our automatically constructed execution-verified data and RILA are highly complementary. Supervised fine-tuning on our data improves the base model from \textbf{40.40\%} to \textbf{42.78\%}, and applying RILA on top of the fine-tuned model further boosts the final score to \textbf{57.52\%}---a gain of \textbf{+14.74} over its one-shot baseline and \textbf{+17.12} over the original 9B backbone. This shows that better initialization from supervised fine-tuning and execution-driven refinement address different aspects of webpage generation and combine effectively. Our central finding is that this combination lets a compact 9B model outperform far larger models that do not use RILA: guided by a Kimi-K2.6 critique, the fine-tuned Qwen3.5-9B$^\dagger$ surpasses both the 1T-parameter Kimi-K2.6 (\textbf{55.61\%}) and the proprietary GPT-5.5 (\textbf{55.74\%}) under one-shot generation, indicating that execution-driven refinement with a strong critique can compensate for a large gap in code-model scale.

RILA also benefits the stronger models themselves. Under a Kimi-K2.6 critique, GPT-5.5 improves from \textbf{55.74\%} to \textbf{63.00\%}, while Kimi-K2.6 improves from \textbf{55.61\%} to \textbf{65.85\%}, the best result in Table~\ref{tab:main_results}, showing that execution-aware browser feedback remains beneficial even for models that already possess strong code generation capabilities.

\subsection{Ablation Study}

We analyze the key design choices of RILA, including its execution-driven refinement mechanisms (best-so-far selection, execution feedback, and tool-assisted editing), the choice of critique model, and the SWE iteration budget, and we further probe its generalization on Interaction2Code.

\begin{table}[t]
\centering
{\small
\setlength{\tabcolsep}{4pt}
\begin{tabular}{ccccccc}
\toprule
\shortstack{Best-so-\\far} & \shortstack{Exec.\\Feedback} & \shortstack{Tool\\Editing} & \shortstack{Iter.\\Refine.} & IFS $\uparrow$ & VFS $\uparrow$ & Final $\uparrow$\\
\midrule
\checkmark & \checkmark & \checkmark & \checkmark & \textbf{46.90} & \textbf{58.38} & \textbf{50.34}\\
           & \checkmark & \checkmark & \checkmark & 44.49 & 58.05 & 48.56\\
           &            & \checkmark & \checkmark & 44.03 & 54.34 & 47.12\\
           &            &            & \checkmark & 42.47 & 52.63 & 45.52\\
\bottomrule
\end{tabular}}
\caption{
Cumulative component ablation of RILA on the Qwen3.5-9B backbone; a checkmark marks an enabled mechanism (higher is better). Columns abbreviate \textbf{Best-so-far Selection} (carrying the ERS-best candidate into the next round), \textbf{Execution Feedback} (feeding browser-rendered observations to the critique), \textbf{Tool Editing} (localized multi-step edits with SWE tools), and \textbf{Iterative Refinement} (the shared three-round loop). Disabling Tool Editing (last row) instead regenerates the full webpage from the critique each round.
}
\label{tab:best-code-ablation}

\vspace{1em}

\centering
\setlength{\tabcolsep}{4pt}
\begin{tabular}{llccc}
\toprule
Code Model & Critique Model & IFS $\uparrow$ & VFS $\uparrow$ & Final $\uparrow$\\
\midrule
Qwen3.5-9B & Qwen3.5-9B$^\dagger$ & 40.61 & 55.48 & 45.07\\
Qwen3.5-9B & Kimi-K2.6 & 46.90 & 58.38 & \textbf{50.34}\\
\midrule
Qwen3.5-9B$^\dagger$ & Qwen3.5-9B$^\dagger$ & 46.12 & 62.45 & 51.02\\
Qwen3.5-9B$^\dagger$ & Kimi-K2.6 & 53.91 & 65.96 & \textbf{57.52}\\
\bottomrule
\end{tabular}
\caption{
Effect of the critique model in RILA on IWR-Bench. For each code model, we pair it with a compact critique (Qwen3.5-9B$^\dagger$) and a stronger one (Kimi-K2.6). Higher is better.
}
\label{tab:critique_ablation}
\end{table}

\begin{figure}[t]
\centering
\includegraphics[width=0.95\columnwidth]{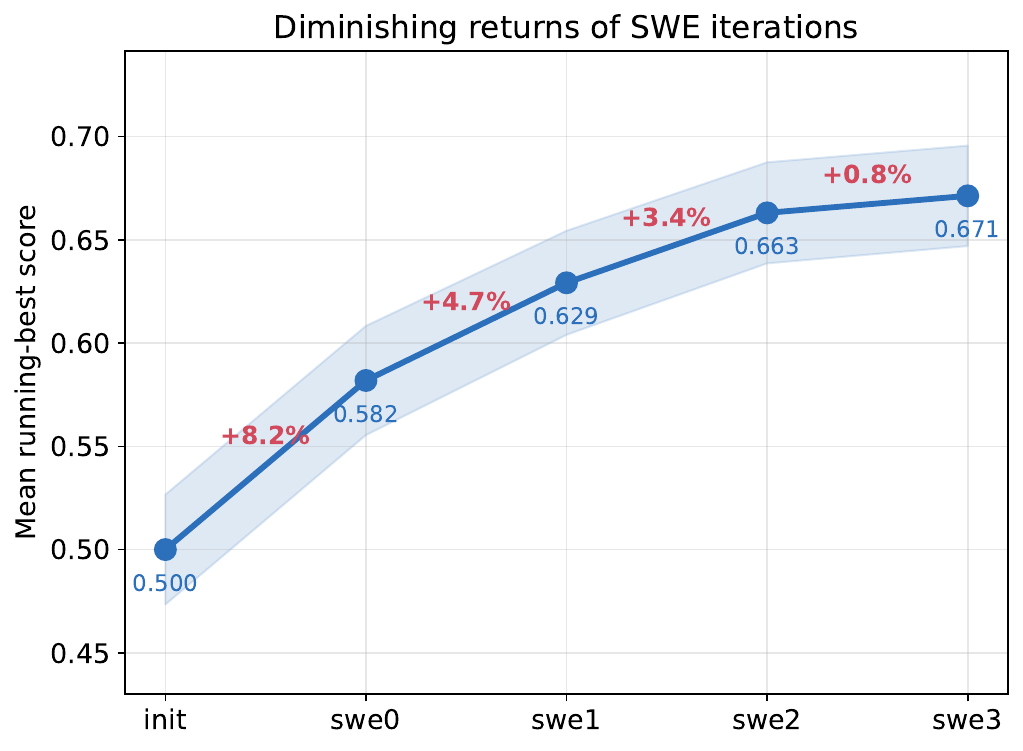}
\caption{
Ablation of the SWE iteration budget on IWR-Bench (113 tasks). We plot the running-best ERS across successive SWE iterations; the marginal gain diminishes rapidly and largely saturates after three iterations.
}
\label{fig:swe_times}
\end{figure}

\textbf{Each mechanism improves complementary metrics.}
Table~\ref{tab:best-code-ablation} reports a cumulative ablation on the Qwen3.5-9B backbone with the budget fixed at three rounds, and the mechanisms improve complementary axes. Removing best-so-far selection lowers Final by 1.78 (50.34$\rightarrow$48.56), almost entirely from a 2.41-point IFS drop (VFS $-0.33$): its interaction-outcome signal prevents a correct interactive implementation from being overwritten by later edits. Further removing execution feedback (which also disables ERS) instead reduces VFS by 3.71 (58.05$\rightarrow$54.34) with little IFS change ($-0.46$), as the rendered screenshots that expose layout and styling defects are no longer available. Finally, replacing the tool-assisted SWE agent with single-shot critique-conditioned regeneration degrades both axes (IFS $-1.56$, VFS $-1.71$; 45.52 Final), since rewriting the whole page discards already-correct regions rather than applying targeted fixes. Together with the diminishing returns of additional SWE iterations (Figure~\ref{fig:swe_times}), effective refinement thus depends jointly on execution-grounded feedback, the conditioning strategy, and the editing mechanism.

\textbf{RILA generalizes across critique models.}
RILA relies on a critique model to turn execution feedback into actionable fixes, so its capability upper-bounds refinement quality. Table~\ref{tab:critique_ablation} varies the critique model while fixing the code model. A stronger critique (Kimi-K2.6) yields consistently larger gains than a compact critique (Qwen3.5-9B$^\dagger$): it raises the Final score from 45.07 to 50.34 for the Qwen3.5-9B code model and from 51.02 to 57.52 for the fine-tuned Qwen3.5-9B$^\dagger$. Notably, even a compact critique---using Qwen3.5-9B$^\dagger$ itself---already improves over one-shot generation (40.40$\rightarrow$45.07 and 42.78$\rightarrow$51.02), showing that RILA does not require a large critique model, while a stronger critique further widens the gain.

\textbf{Effect of the SWE iteration budget.}
We extend the refinement budget to four iterations on IWR-Bench and record the running-best ERS at each step. As shown in Figure~\ref{fig:swe_times}, it increases monotonically from 0.500 to 0.671 while the marginal gain rapidly diminishes: the fourth iteration adds only about one-tenth of the first, and the first three already recover 95.3\% of the total improvement. We therefore adopt three SWE iterations throughout.

\textbf{Cross-benchmark generalization.}
To test whether RILA's refinement transfers beyond IWR-Bench, we evaluate the same agent unchanged on Interaction2Code~\citep{xiao2024interaction2code}, an independently collected benchmark for reproducing interactive webpages from screenshots, using Kimi-K2.6 to guide the Qwen3.5-9B code model. Unlike IWR-Bench's continuous multi-step trajectories, Interaction2Code annotates each page with mutually independent single-step transitions (a source state, a trigger, and the resulting destination); we therefore decompose every page into single-interaction units and cast each into our interaction schema, preserving the original annotation semantics while matching a per-interaction refinement protocol, which yields 375 tasks. We score with the official Interaction2Code evaluator, which reports CLIP, SSIM, and text similarity over the full page and the localized interaction region, together with an interaction position-agreement score and an interaction implementation rate. As shown in Table~\ref{tab:i2c_generalization}, enabling refinement improves every metric, with the largest gains on interaction implementation ($+0.027$) and text fidelity (full page $+0.025$, interaction region $+0.021$), together with a clear gain in interaction localization (position $+0.019$). RILA thus remains effective under a different input modality, a different interaction format, and a different evaluator, indicating that its execution-driven refinement is not specific to IWR-Bench.

\begin{table}[t]
\centering
\small
\setlength{\tabcolsep}{5pt}
\begin{tabular}{llccc}
\toprule
View & Metric & w/o RILA & w/ RILA & $\Delta$ \\
\midrule
\multirow{3}{*}{Full page}
 & CLIP $\uparrow$ & 0.821 & \textbf{0.839} & +0.018 \\
 & SSIM $\uparrow$ & 0.697 & \textbf{0.707} & +0.010 \\
 & Text $\uparrow$ & 0.766 & \textbf{0.791} & +0.025 \\
\midrule
\multirow{5}{*}{Interaction}
 & CLIP $\uparrow$     & 0.728 & \textbf{0.739} & +0.011 \\
 & SSIM $\uparrow$     & 0.550 & \textbf{0.554} & +0.004 \\
 & Text $\uparrow$     & 0.429 & \textbf{0.451} & +0.021 \\
 & Position $\uparrow$ & 0.441 & \textbf{0.460} & +0.019 \\
 & IR $\uparrow$       & 0.851 & \textbf{0.877} & +0.027 \\
\bottomrule
\end{tabular}
\caption{Generalization to Interaction2Code. IR (interaction implementation rate) is computed over all 375 tasks; the remaining similarity metrics are averaged over the 301 tasks whose interaction is triggered by \emph{both} variants, giving a strictly paired comparison. Higher is better.}
\label{tab:i2c_generalization}
\end{table}

\section{Conclusion}

We presented \textbf{RILA}, an execution-driven agent for multimodal web development that runs each generated webpage in a real browser and iteratively improves it through an execution--critique--edit loop. An \textbf{Action Interaction Verification (AIV)} module replays the reference trajectory and verifies every action, an \textbf{Execution-Aware Rendering Score (ERS)} jointly measures interaction correctness and visual fidelity and retains the historically best implementation, and a multimodal critique turns these observations into structured fixes applied with tool-assisted software engineering. We further introduced an execution-verified data construction pipeline that synthesizes diverse, high-quality training data with gains complementary to inference-time refinement. On IWR-Bench, RILA consistently improves both interaction correctness and visual fidelity across backbones; guided by a strong critique, it even lets a compact 9B model surpass far larger open and proprietary models that generate webpages in one shot. By keeping rendering in the loop, RILA turns runtime interaction into a verifiable optimization signal for reliably improving webpage quality, and we hope it offers a foundation for future execution-driven coding agents.

\bigskip

\bibliography{aaai2027}


\end{document}